\documentclass[10pt,twocolumn,letterpaper]{article}

\usepackage{cvpr}
\usepackage{times}
\usepackage{epsfig}
\usepackage{graphicx}
\usepackage{amsmath}
\usepackage{amssymb}
\usepackage{booktabs}
\usepackage{multirow}
\usepackage{xcolor}
\usepackage{cite}
\definecolor{citecolor}{HTML}{0071bc}

\usepackage[pagebackref=true,breaklinks=true,letterpaper=true,colorlinks,citecolor=citecolor,bookmarks=false]{hyperref}

\newcommand\jc[1]{{#1}}

\cvprfinalcopy 

\ifcvprfinal\fi
\begin{document}

\title{The Semi-Supervised iNaturalist Challenge at the FGVC8 Workshop}

\author{Jong-Chyi Su \quad \quad  Subhransu Maji\\
University of Massachusetts Amherst\\
{\tt\small \{jcsu, smaji\}@cs.umass.edu}
}

\maketitle
\thispagestyle{plain}
\pagestyle{plain}

\begin{abstract}
Semi-iNat is a challenging dataset for semi-supervised classification with a long-tailed distribution of classes, fine-grained categories, and domain shifts between labeled and unlabeled data.
This dataset is behind the second iteration of the semi-supervised recognition challenge to be held at the FGVC8 workshop at CVPR 2021.
Different from the previous one, this dataset (i) includes images of species from different kingdoms in the natural taxonomy, (ii) is at a larger scale --- with 810 in-class and \jc{1629} out-of-class species for a total of $\approx$330k images, and (iii) does not provide in/out-of-class labels, but provides coarse taxonomic labels (kingdom and phylum) for the unlabeled images.
This document describes baseline results and the details of the dataset which is available here: \url{https://github.com/cvl-umass/semi-inat-2021}.

\end{abstract}

\section{Introduction}


Semi-supervised image classification concerns the problem of learning a classifier in the presence of labeled and unlabeled images. While there has been a considerable number of techniques proposed in recent years, progress has been limited in part due to the lack of realistic benchmarks and evaluation protocols. To this end, we presented a semi-supervised dataset called Semi-Aves~\cite{su2021semisupervised} as part of the FGVC7 workshop.
\jc{Semi-Aves contains 1000 bird species where labeled data are from 200 species. Both labeled and unlabeled data have long-tailed distributions.}
The challenge exposed some limitations of existing methods, such as the lack of robustness of out-of-domain unlabeled data and limited improvements in the transfer learning setting~\cite{su2021realistic}.
Motivated by this we present another iteration of this challenge to be held at the FGVC8 workshop at CVPR 2021.

\begin{figure}[t!]
\newcommand{\w}{0.23}
\newcommand{\h}{0.12}
\centering
\setlength\tabcolsep{0.5pt}
\begin{tabular}{cccc}
\small{Mollusca} & \small{Chordata}& \small{Arthropoda}& \small{Echinodermata}\\
\includegraphics[width=\w\linewidth,height=\h\textwidth,trim={0cm 1cm 0cm 1cm},clip]{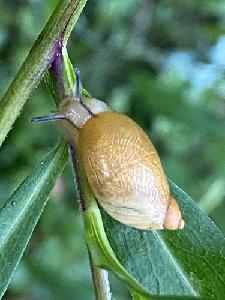}&
\includegraphics[width=\w\linewidth,height=\h\textwidth,trim={1.5cm 0cm 1.5cm 0},clip]{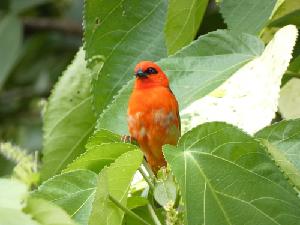}&
\includegraphics[width=\w\linewidth,height=\h\textwidth]{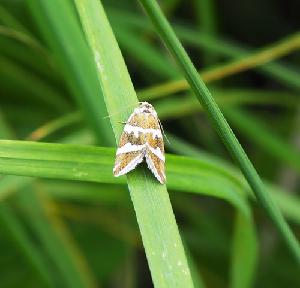}&
\includegraphics[width=\w\linewidth,height=\h\textwidth,trim={1cm 0cm 1cm 0},clip]{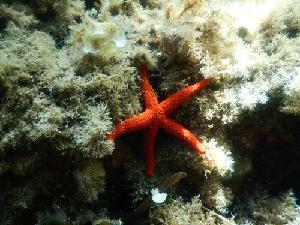}\\
\small{Tracheophyta}&  \small{Bryophyta}&  \small{Basidiomycota}&  \small{Ascomycota}\\
\includegraphics[width=\w\linewidth,height=\h\textwidth]{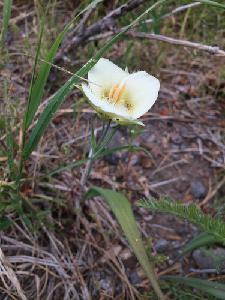}&
\includegraphics[width=\w\linewidth,height=\h\textwidth]{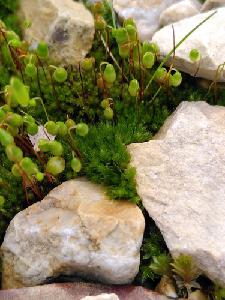}&
\includegraphics[width=\w\linewidth,height=\h\textwidth]{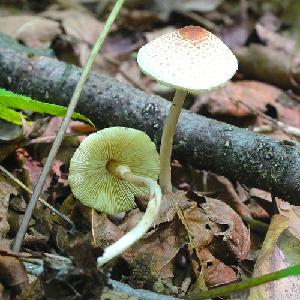}&
\includegraphics[width=\w\linewidth,height=\h\textwidth]{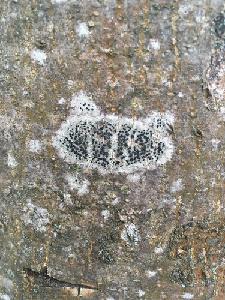}\\
\end{tabular}
\caption{\textbf{Examples from the Semi-iNat dataset.} The Semi-iNat dataset includes images from 8 different phyla.}
\label{fig:images}
\end{figure}

Semi-iNat builds on the previous year's benchmark while incorporating some new challenges.
First, the dataset contains species from three kingdoms: Animal, Plants, and Fungi (Fig.~\ref{fig:images} and Tab.~\ref{tab:taxa}), unlike Semi-Aves which contains only Aves (birds). 
There are a total of 2439 animal species and $\approx$330k images, which is \jc{more than two times} larger than Semi-Aves.
In addition, we do not provide the label if the unlabeled images are from the same classes as labeled data (which we did in Semi-Aves), and instead provide the coarse taxonomy for unlabeled data.
Coarse taxonomic labels are easily obtained from non-experts and provide a weak supervisory signal that could be exploited by the learning methods.
In the following sections, we describe the data collection process and provide baseline performances.

\begin{table}[t!]
\setlength{\tabcolsep}{7pt}
\renewcommand{\arraystretch}{1.2}
\centering
\begin{tabular}{c c c c c}
\toprule
\textbf{Kingdom}&	\textbf{Phylum}&	    \textbf{$C_{in}$}&	\textbf{$C_{out}$}\\
\midrule
\multirow{4}{*}{\shortstack{Animalia\\(1294)}}&   Mollusca&   11 &    24\\
&     Chordata&   113 &   228\\
&    Arthropoda& 301 &   605\\
&    Echinodermata& 4 &    8\\
\midrule
Plantae&    Tracheophyta&  336& 674\\
(1028)&     Bryophyta& 6&   12\\
\midrule
Fungi&      Basidiomycota& 29&  58\\
(117)&      Ascomycota& 10&    20\\
\bottomrule
\end{tabular}
\vspace{0.05in}
\caption{\textbf{The number of species in the taxonomy.} For each phylum, we select around one-third of the species for the in-class set $C_{in}$ and the rest for the out-of-class set $C_{out}$.}
\label{tab:taxa}
\end{table}

\begin{table*}[h!]
\setlength{\tabcolsep}{7pt}
\renewcommand{\arraystretch}{1.3}
\centering
\begin{tabular}{c c c c c | c c c c c}
\toprule
\multicolumn{5}{c|}{\textbf{Semi-Aves}} & \multicolumn{5}{c}{\textbf{Semi-iNat}}\\
\textbf{{Split}}&	\textbf{{Details}}&	\textbf{{Classes}}& \textbf{{\#Images}} & \textbf{\#img/cls} & \textbf{{Split}}&	\textbf{{Details}}&	\textbf{{Classes}}& \textbf{{\#Images}} & \textbf{\#img/cls}\\
\midrule
Train&	Labeled&	200&	3,959 & 5-43 & Train&	Labeled&	810&	9,721 & 5-80\\
Train&	Unlabeled (in)&	200&	26,640 & 16-229 & \multirow{2}{*}{Train}&	\multirow{2}{*}{Unlabeled}&	\multirow{2}{*}{2439}&	\multirow{2}{*}{313,248} & \multirow{2}{*}{19-400} \\
Train&	Unlabeled (out)&	800&	122,208 & 23-250 &&&&&\\
Val&	Labeled&	200&	2,000 & 10 & Val&	Labeled&	810&	4,050 & 5 \\
Test&	Public&	    200&	4,000 & 20 & Test&	Public&	    810&	8,100 & 10 \\
Test&	Private&	200&	4,000 & 20 & Test&	Private&	810&	8,100 & 10 \\
\bottomrule
\end{tabular}
\vspace{0.05in}
\caption{
\textbf{Statistics of the Semi-Aves~\cite{su2021semisupervised} and Semi-iNat dataset.} Both labeled and unlabeled training sets have a long-tailed distribution, while validation and test sets are uniformly distributed. The test set splits are only used for the Kaggle competition at the FGVC8 workshop. 
\jc{We also show the statistics of Semi-Aves~\cite{su2021semisupervised} on the left for a comparison. Different from Semi-Aves, Semi-iNat has more species and images, and a single set of unlabeled data.}
}
\label{tab:stats}
\end{table*}

\section{The Semi-iNat Dataset}

We create the dataset including organisms across different kingdoms in the taxonomy rank. 
We first remove the species that appear in the previous iNat challenges~\cite{van2018inaturalist}, including iNat-17, iNat-18, iNat-19, iNat-21, and the Semi-Aves~\cite{su2021semisupervised} challenges, to avoid any overlap.
We then select the most common species from the iNaturalist website~\cite{iNaturalist}. 
To ensure the diversity of images in each category, we only select one image per contributor (photographer). 

\begin{figure}[t!]
\centering
\includegraphics[width=\linewidth]{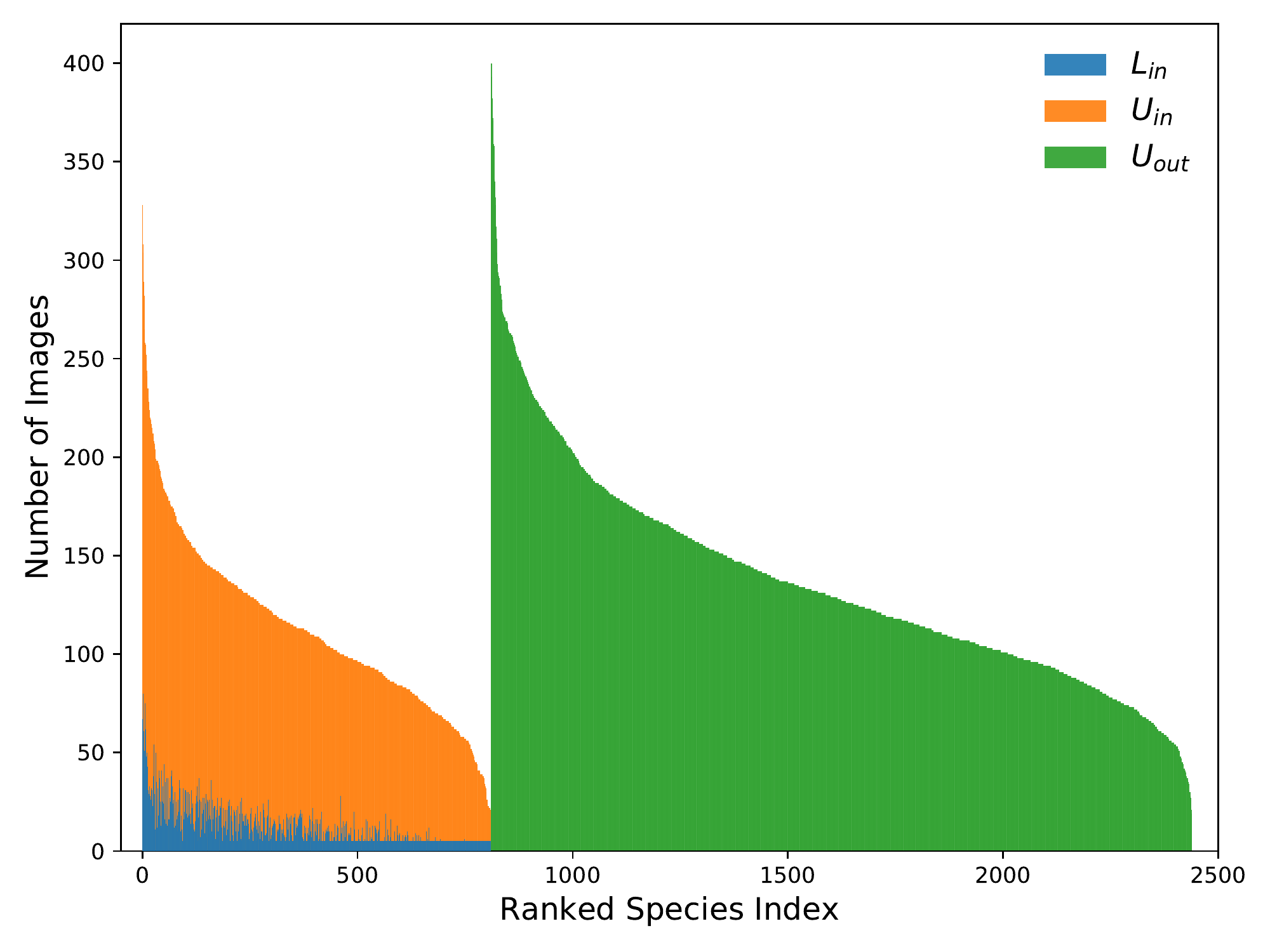}
\caption{\textbf{Class distribution of Semi-iNat dataset.} Note that the distinction between $U_{in}$ and $U_{out}$ are only shown here for visualization. We did not provide the domain labels for unlabeled data.}
\label{fig:hist}
\end{figure}

Next, we split the species into two sets $C_{in}$ and $C_{out}$, one for in-class and one for out-of-class data. To avoid a large domain mismatch between them, we make the split depending on the taxonomy. For each phylum, we randomly select about one-third of the species for $C_{in}$, and the rest for $C_{out}$. The number of species in the taxonomy is shown in Tab.~\ref{tab:taxa}. 
For each species in $C_{in}$, we first select 5/10/10 images for validation, public test, and private test set. Among the rest of the images, we then sample around 10\% of the images as labeled data $L_{in}$ and the rest as unlabeled data $U_{in}$. 
We also ensure a minimum of 5 labeled images per class.
For species in $C_{out}$, there are up to 400 images and all of them are included in $U_{out}$. The two sets of unlabeled data $U = U_{in} \cup U_{out}$ are then combined and no domain labels are provided.
The statistics of the class distribution are shown in Tab.~\ref{tab:stats} and the histogram is shown in Fig.~\ref{fig:hist}.


In addition to class labels, we also provide the full taxonomy of the species in $C_{in}$. For each unlabeled image in $U_{out}$, we provide its kingdom and phylum information. 
This is to model a scenario where coarse labels can be acquired much more easily than fine-grained ones and may be used to improve the performance of SSL algorithms.
\jc{Last, we restrict the image resolution to be at most 300px on either side, different from Semi-Aves where its at most 500px.}

\section{Baselines}

We present results using ResNet-50~\cite{he2016deep} models trained from scratch or from ImageNet pre-trained model.
We also train models using the labels $L_{in}$ only (baseline), and when including the ground-truth labels of images from $U_{in}$ (oracle).
We train using stochastic gradient descent with cosine learning rate decay and a batch size of 64 following~\cite{su2021realistic}. 
The learning rate and weight decay are tuned between [1e-2, 3e-3] and [1e-3 1e-5]. We train the model for 50k and 150k iterations when ImageNet pre-training is used. For training from scratch, we use 3$\times$ more iterations.
Tab.~\ref{tab:result} shows that ImageNet pre-training provides a 22\% improvement over training from scratch. The gap between baseline and oracle shows that there is ample room for improvement. 

\jc{
We further analyze the predictions of the baseline model (trained from ImageNet) by showing the confusion matrix. We group the categories on the phylum level and plot the normalized 8$\times$8 confusion matrix in Fig.~\ref{fig:confusion}. 
The diagonal shows that the accuracy on each phylum is proportional to the number of species and images. 
On off-diagonal entries, we can see more confusions within the kingdom. 
}

\begin{table}[t!]
  \setlength{\tabcolsep}{9.5pt}
  \renewcommand{\arraystretch}{1.2}
  \centering
  \begin{tabular}{c | c c | c c }
    \toprule
    \multirow{2}{*}{Method} & \multicolumn{2}{c|}{\textbf{from scratch}} & \multicolumn{2}{c}{\textbf{from ImageNet}} \\
     & Top-1 & Top-5 & Top-1 & Top-5 \\
    \midrule
    Baseline & 19.2\%&	36.4\%&	41.0\%&	63.7\%\\
    Oracle   & 93.3\%&	96.9\%&	94.3\%&	97.5\%\\
    \bottomrule
    \end{tabular}
    \vspace{0.05in}
    \caption{\textbf{Performance of ResNet-50 on Semi-iNat.} The baseline is trained on labeled data only, while the oracle is trained on all the data and their (hidden) labels in $L_{in} \cup U_{in}$.}
    \label{tab:result}
\end{table}

\begin{figure}[t!]
\setlength{\tabcolsep}{9.5pt}
\renewcommand{\arraystretch}{1.2}
\centering
\includegraphics[width=\linewidth]{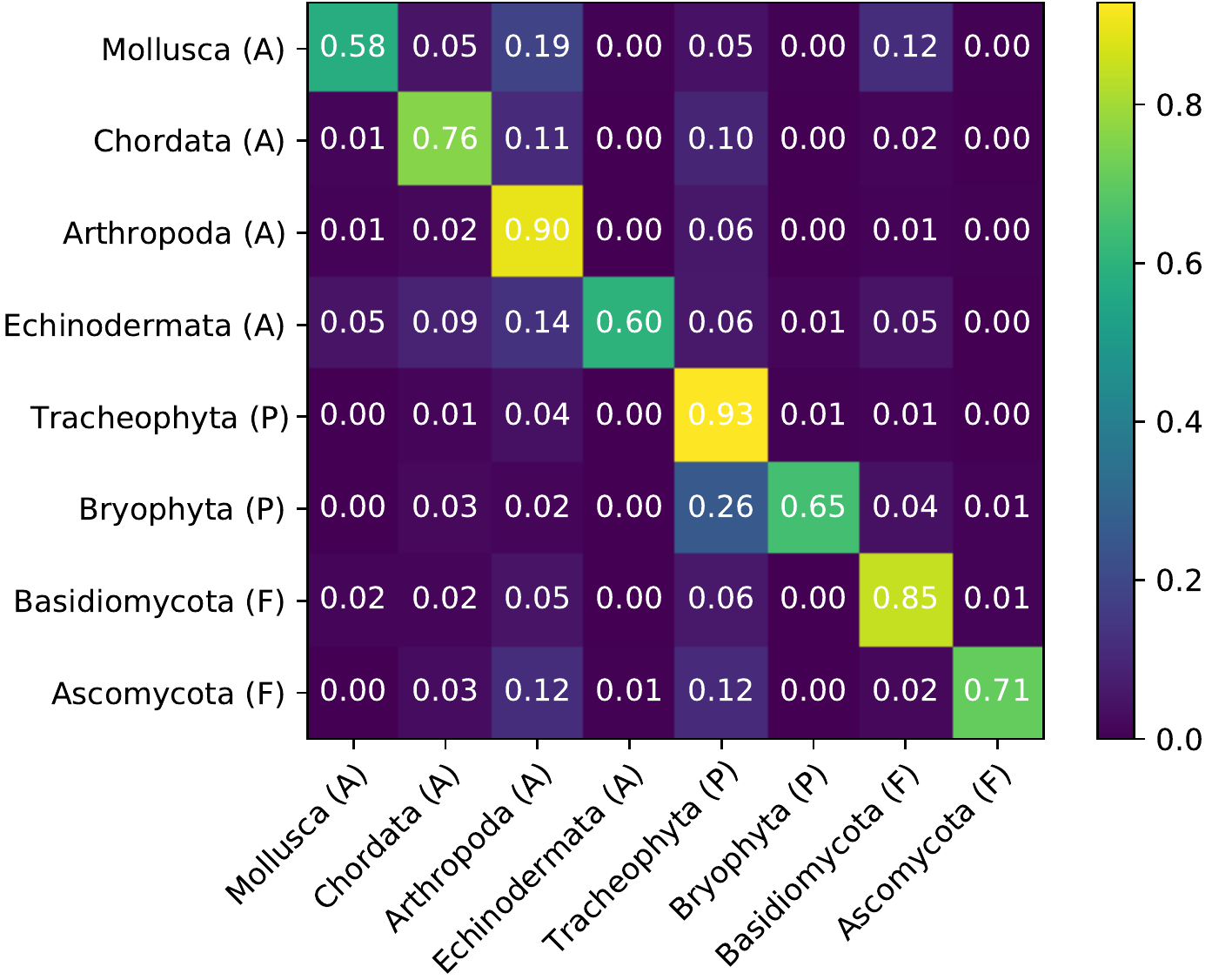}
\caption{
\jc{
\textbf{Confusion matrix of the baseline model on the Phylum level.} The kingdoms are shown in the brackets: (A) for Animalia, (P) for Plantae, and (F) for Fungi. We can see that the accuracy is lower for those phylum with less species, and more confusions come from the species under the same kingdom.
}
}
\label{fig:confusion}
\end{figure}

\paragraph{Acknowlegements.} This project is supported in part by NSF \#1749833. We also thank the FGVC team and Kaggle for organizing the workshop; Oisin Mac Aodha and Grant van Horn for the help collecting the dataset.

{\small
\bibliographystyle{ieee_fullname}
\bibliography{egbib}
}

\end{document}